\documentclass[review,12pt,3p]{elsarticle}

\usepackage[utf8]{inputenc}
\usepackage[T1]{fontenc}

\usepackage{lineno,hyperref}
\usepackage{float}
\usepackage{enumerate}
\usepackage{amsmath}
\usepackage{multirow}
\usepackage{booktabs}
\usepackage{graphicx}
\usepackage{verbatim}
\usepackage[dvipsnames]{xcolor}
\usepackage{subcaption}
\usepackage{amssymb}
\usepackage{pifont}
\usepackage{tablefootnote}
\usepackage{lscape}

\newcommand{\cmark}{\ding{51}}%
\newcommand{\xmark}{\ding{55}}%

\modulolinenumbers[5]

\journal{Journal of \LaTeX\ Templates}

\begin{document}

\begin{abstract}
Hate speech is a major issue in social networks due to the high volume of data generated daily. Recent works demonstrate the usefulness of machine learning (ML) in dealing with the nuances required to distinguish between hateful posts from just sarcasm or offensive language. Many ML solutions for hate speech detection have been proposed by either changing how features are extracted from the text or the classification algorithm employed. However, most works consider only one type of feature extraction and classification algorithm. This work argues that a combination of multiple feature extraction techniques and different classification models is needed. We propose a framework to analyze the relationship between multiple feature extraction and classification techniques to understand how they complement each other. The framework is used to select a subset of complementary techniques to compose a robust multiple classifiers system (MCS) for hate speech detection. The experimental study considering four hate speech classification datasets demonstrates that the proposed framework is a promising methodology for analyzing and designing high-performing MCS for this task.
MCS system obtained using the proposed framework significantly outperforms the combination of all models and the homogeneous and heterogeneous selection heuristics, demonstrating the importance of having a proper selection scheme. Source code, figures and dataset splits can be found in the GitHub repository: \url{https://github.com/Menelau/Hate-Speech-MCS}.

%achieving state-of-the-art results in public datasets. Results also show the importance of considering multiple representations and multiple classification models to improve hate speech detection performance. Source code and dataset splits can be found in the GitHub repository: \url{https://github.com/Menelau/Hate-Speech-MCS}.

\end{abstract}


\begin{frontmatter}

    \title{Selecting and combining complementary feature representations and classifiers for hate speech detection}
    %Hate speech detection using classifier combination}
    %\author{Rafael M. O. Cruz, Woshington V. de Sousa, George D. C. Cavalcanti}
    %\address{}
    
    \author[ets]{Rafael M. O. Cruz\corref{corr1}}
    \ead{rafael.menelau-cruz@etsmtl.ca}
    
    \author[ufpe]{Woshington V. de Sousa}
    \ead{wvs2@cin.ufpe.br}
    
    \author[ufpe]{George D. C. Cavalcanti}
    \ead{gdcc@cin.ufpe.br}
    					
    \address[ets]{LIVIA, \'{E}cole de Technologie Sup\'{e}rieure, University of Quebec, Montreal, Qu\'{e}bec, Canada - www.liviamtl.ca}
    
    \address[ufpe]{Centro de Inform\'{a}tica, Universidade Federal de Pernambuco, Recife, PE, Brazil\\}% - www.cin.ufpe.br/$\sim$viisar\\}
    					
    \cortext[corr1]{Corresponding author. Email Address: rafael.menelau-cruz@etsmtl.ca}
\end{frontmatter}

\section{Introduction}

Social networks are an essential communication channel with an exponential increase in the last years. It is considered relevant in several contexts because it allows users to express opinions, share content, and disseminate news. They impact millions of users every day and change the general socialization perspective~\cite{batool2012}.

Social media allows users to register quickly and propagate content from the most diverse subjects~\cite{ensembleFauzi2018}. Due to a large amount of content published daily on these platforms, the task of monitoring and control its contents is arduous. The lack of proper moderation provides freedom for users to perform cyberbullying or other forms of attacks. The dissemination of hate speech becomes very common in this scenario since anybody with an Internet connection can become a disseminating agent. Users can also use creating fake profiles to avoid identification, thus becoming more confident that they will not be punished.
Another problem is bots, which are programmed algorithms that behave like humans in performing tasks like disseminating news and information. Therefore, it can also act as disseminating hate speech  agent~\cite{howard2016bots}. According to the data provided by Twitter, approximately 8.5\% of users in the platform are bots~\cite{botsTwitter}. 

Hate speech often appears in incidents with a tremendous social repercussion such as the American presidential election, Brexit~\cite{trumpAndBRexit}, terrorist attacks associated with Islamic groups~\cite{scharwachter2020does} and the COVID-19 pandemic, that provoked discrimination against Chinese and Asians \cite{hu2020covid}.

\begin{figure}[htb]
	\centering
	\includegraphics[height=8cm]{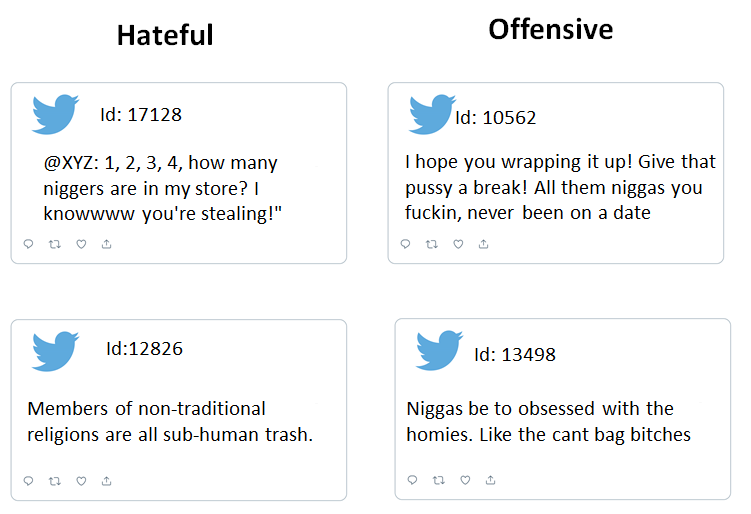}
	\caption{Examples of hateful and offensive tweets extracted from the Thomas Davidson dataset~\cite{Hate-Speech-Offensive}. The ID tag indicate their corresponding unique ids in the full dataset.}
	\label{fig:hatevsoffensive}
\end{figure}

Detecting hate speech is a complex task because it can be easily mistaken with offensive language and humor posts that, in many situations, are protected by laws of freedom of speech, as the Brazilian Civil Rights Framework for the Internet, which was established in 2014. 
Hate speech is usually defined as any sort of speech targeting a particular group, especially based on race, religion, or sexual orientation. Davidson et al.~\cite{DWMW17} defined hate speech as: ``{\it Language that is used to express hatred towards a targeted group or intended to be derogatory, to humiliate, or to insult the members of the group}''. In contrast, offensive language is characterized by the use of terms that are highly offensive to some groups, such as \emph{n*gga}, and \emph{f*g}, which are prevalent on social media. However, the text itself is not targeting or inciting violence to a specific group. Figure~\ref{fig:hatevsoffensive} illustrates two examples of hateful and two offensive tweets extracted from the Thomas Davidson dataset~\cite{Hate-Speech-Offensive}. This nuance makes human moderation a slow and challenging process. Hence, impossible to deal with a large amount of content created daily.

\textcolor{blue}{In this paper, we explore the problem of automatic hate speech detection in English posts from social media.} Two crucial aspects of creating a machine learning solution are extracting features from the text and the classification algorithm employed~\cite{schmidt2017survey}. There are several feature extraction techniques such as Bag of Word approaches~\cite{soumya2014text}, lexical resources~\cite{gitari2015lexicon}, sentiment analysis~\cite{van2015detection} and dense word vectors~\cite{glove2014, mikolov2013}. The classification algorithm can be classical methods (e.g., Logistic Regression and Naïve Bayes) or Deep Learning approaches like Convolutional Neural Networks (CNN)~\cite{zhang2018}. Multiple works in hate speech detection are now using an ensemble model to achieve higher performance. However, comparative studies showing which are the most effective feature extraction techniques and classification methods for this problem is still an open research question~\cite{schmidt2017survey}. Furthermore, the field lacks a methodology to analyze multiple feature representation techniques and classification algorithms.

We propose a framework based on classification diversity to analyze different feature representation methods' behavior and their impact on multiple classification algorithms. The framework allows us to select a set of complementary feature representation and classifier models to obtain a Multiple Classifier System (MCS) that significantly improves hate speech detection. The framework also allows evaluating the impact of changing the feature extraction method and the classification model to know which component has a more significant impact on hate speech detection performance. 

\textcolor{blue}{We considered four English hate-speech detection datasets in this study.} Results demonstrate that combining all models without any proper selection scheme produces sub-optimal results. Selecting a suitable combination of feature extraction techniques and classification algorithms achieves state-of-the-art performance for multiple datasets, demonstrating the importance of carefully selecting a complementary set of models for improving performance. 

Hence, the contributions of this work are:

\begin{enumerate}
	\item A novel framework to analyze the relationship between multiple feature representation techniques and classification algorithms.
	
	\item The evaluation of different strategies to compose an MCS and their interpretation based on the proposed framework.
	
	\item A Multiple Classifier System with a diverse set of feature representation and classification models to achieve state-of-the-art results for this task.
\end{enumerate}

This paper is organized as follows: Section~\ref{sec:related_works} introduces hate speech detection problems and discusses the related works involving machine learning. Section~\ref{sec:approach} presents the proposed multi-view hate speech detection framework. Section~\ref{sec:methodology} describes the datasets and the experimental methodology used in this study. Section~\ref{sec:sensitivity} presents an analysis of multiple classifier models and feature extraction techniques to build a high-performance MCS for the hate speech detection problem. Experimental results and their analyses are conducted in Section~\ref{sec:results}. \textcolor{blue}{Conclusion and future works are presented in Section~\ref{sec:conclusion}.}

\section{Related work}
\label{sec:related_works}

Supervised learning requires each instance's association to a label for a machine learning algorithm's training process. The hate speech detection task is cast as a supervised one, and, regarding the classification phase, it can be divided into two approaches: monolithic or ensemble. In the former, only one model is employed, while, in the latter, many models are trained and combined to output the predicted class of query instances.

Table~\ref{tab:related_works} shows literature works that address hate speech using monolithic classifiers and ensemble learning. Recent papers use word embedding methods more frequently than bag-of-words and n-grans because the former can extract semantic information from the text; consequently, an improvement in the performance is expected.
Regarding the classifier, different paradigms have been employed; tree-based algorithms such as decision trees and random forest (RF)~\cite{BW15,DWMW17,MS18,RZW19}, artificial neural networks such as multi-layer perceptron (MLP) and convolution neural networks (CNN)~\cite{PF17,ZK18,PRL18,zhang2018,PSD20, ZYH20, ASK20, ASM20, BRB20, PGLV21,Jan21}, Bayesian as the naive bayes (NB)~\cite{BW15,DWMW17}, support vector machines (SVM)~\cite{BW15,DWMW17}, and ensemble learning, which is marked (\cmark) in the last column of the table.
%~\cite{BW15,ZK18,PRL18,HBAW19,AT20}. 

\begin{landscape}
	\begin{table}[H]
		\centering
		\scalebox{0.6}{
			\begin{tabular}{lcccccc}
				\hline
				Year & Citation & Dataset & Dataset Category &Feature & Classifier & Ensemble \\ \hline
				
				2015 & Burnap and Williams~\cite{BW15} & 1,901 tweets & race ethnicity, religion & BoW and n-gram & BLR, RF, SVM & \cmark \\ \hline 
				%& & (11.68\% of hate and 88.32\% of non-hate) & & \\ \hline
				
				2017 & Davidson et al.~\cite{DWMW17} &  24,802 tweets & racism, sexism, homophobia & n-gram &  LR, NB, DT, RF, SVM & \xmark \\ \hline  
				
				2017 & Park and Fung~\cite{PF17} & Twitter from~\cite{WH16,Was16}  & racism, sexism &  characters, words, and both & CNN & \xmark \\ \hline 
				
				2018 & Zimmerman et al.~\cite{ZK18} & Twitter from~\cite{WH16} & racism, sexism & embedding & deep learning & \cmark \\ \hline
				
				2018 & Pitsilis et al.~\cite{PRL18} & Twitter from~\cite{WH16} & racism, sexism & defined by the authors & LSTM & \cmark \\ \hline
				
				2018 & Montani and Schuller~\cite{MS18} & GermEval 2018\tablefootnote{\url{https://github.com/uds-lsv/GermEval-2018-Data}} & general & TFIDF, Word2Vec, n-gram & LR, RF, ET & \cmark \\ \hline
				
				2019 & Zhang and Luo~\cite{zhang2018} &  Twitter from~\cite{BW15, WH16} & ~\cite{BW15}:~race~ethnicity,~religion \cite{WH16}:~racism,~sexism & Word2Vec & CNN & \xmark \\ \hline  
				
				2019 & Liu et al.~\cite{HBAW19} & Twitter from~\cite{BW15} & race ethnicity, religion & embedding, LDA & fuzzy ensemble & \cmark \\ \hline
				
				2019 & Ramakrishnan et al.~\cite{RZW19} & OffensEval~\cite{zampieri-etal-2019-semeval} & general & n-gram, GloVe, others & LR, RF, XG & \cmark \\ \hline
				
				2020 & Paschalides et al.~\cite{PSD20} & Twitter from~\cite{DWMW17} & racism, sexism, homophobia & n-gram, TFIDF, PoS tags, others & CNN, RNN & \cmark \\ \hline
				
				2020 & Al-Makhadmeh and Tolba~\cite{AT20} & Twitter & general & four different features: & deep learning & \cmark \\ 
				& & Stormfront: 10,568 tweets & & semantic, sentiment-based, & & \\
				& & CrowdFlower: 24,783 tweets  & &  unigram, and pattern features & & \\ \hline
				
				2020 & Zhou et al.~\cite{ZYH20} &  HatEval\tablefootnote{https://competitions.codalab.org/competitions/19935}~\cite{Bas19} & misogyny, xenophobia & ELMo, BERT & CNN & \cmark \\ \hline
				
				2020 & Alonso et al.~\cite{ASK20} & HASOC benchmark\tablefootnote{\url{https://hasocfire.github.io/hasoc/2019/}} & general & RoBERTA & CNN & \cmark \\ \hline
				
				2020 & Alsafari et al.~\cite{ASM20} & Arabic-Twitter dataset & general & Fasttext-SkipGram, MBERT, AraBERT & CNN, BiLSTM & \cmark \\ \hline
				
				2020 & Banerjee et al.~\cite{BRB20} & Hindi-English Twitter & general & BERT, RoBERTA, XLNet, DistilBERT & CNN & \cmark \\ \hline
				
				2021 & Plaza-del-Arco et al.~\cite{PGLV21} & HaterNet\tablefootnote{\url{https://zenodo.org/record/2592149\#.YAHX8S271-V}} and HatEval~\cite{Bas19} & HaterNet:~general  \cite{Bas19}:~misogyny,~xenophobia & BERT, XLM, BETO & LSTM, BiLSTM, CNN & \xmark \\ \hline
				
				2021 & Jain et al.~\cite{Jan21} & Kaggle\tablefootnote{\url{www.kaggle.com/arkhoshghalb/twitter-sentiment-analysis-hatred-speech\#train.csv}} & racist, sexist & Word2Vec, GloVe, fastText, ELMo & MLP, BiLSTM & \xmark \\ \hline 
			\end{tabular}
		}
		\caption{Related works. BoW: Bag of Words. BLR: Bayesian Logistic Regression. DT: Decision Tree. ET: ExtraTrees. LDA: Latent Dirichlet Allocation. LR: logistic regression. LSTM: Long Short-Term Memory. NB: Naive Bayes. RF: Random Forest. SVM: Support Vector Machine. XG: XGBoost.}
		\label{tab:related_works}
	\end{table}
\end{landscape}

The most common social media used to extract information to compose a dataset for hate speech detection is Twitter. %, YouTube, Reddit, Facebook, and Wikipedia. 
Despite English being the most used language, there are datasets from many other languages, such as the Arabic-Twitter dataset~\cite{ASM20} and Hindi-English Twitter dataset~\cite{BRB20}.
However, independently of the language, the construction of such datasets is a time-consuming task, not only because the number of hateful instances in online communities is relatively low compared to regular instances, but also because of the labeling process, i.e., which indicates if a sentence is hateful or not, is not a trivial piece of work. 
The labeling process is of utmost importance since supervised learning algorithms use the labels to extract relevant information from the data. Waseem~\cite{Was16} evaluated his own hate speech detection dataset (provided in~\cite{WH16}) to examine the influence of the annotator. He demonstrated that amateur annotators are more likely than expert annotators to label items as hate speech. Moreover, based on his results, systems trained on expert annotations outperform systems trained on amateur annotations.

\subsection{Ensemble for hate speech detection}

Ensemble learning, or multiple classifier systems~(MCS), has proven its effectiveness compared with many monolithic classifiers~\cite{wozniak14,cruz2018dynamic}. MCS are composed of three main parts: generation, selection, and integration~\cite{cruz2018dynamic}. In the first part, {\it generation}, the pool of classifiers is trained, and it can be {\it homogeneous} (all the classifiers use the same learning algorithm, e.g., a pool composed of hundred decision trees) or {\it heterogeneous}, where different learning algorithms are adopted. The second part is optional, and it aims at selecting the best classifiers to compose the final pool. All the selected classifiers' outputs are combined to predict the query pattern class in the integration part. 

Different approaches can be used to integrate the classifier's output. Regarding the ensemble papers listed in Table~\ref{tab:related_works}, the integration methods most used are static rules (such as majority vote, mean, max, mean, and product) and stacked generalization~\cite{wolpert1992stacked}. The latter is a trainable strategy, in opposition to non-trainable strategies such as the static rules, that generates a model that learns the best combination of the pool models' predictions.

Burnap and Williams~\cite{BW15} presented a supervised machine learning text classifier to identify hate speech. They extracted 450,000 tweets in the immediate aftermath of Lee Rigby’s murder. From this amount, a total of 2,000 were labelled by humans and after some preprocessing, only 1,901 tweets (11.68\% of hate and 88.32\% of non-hate) were used. Each tweet was represented using BoW and n-Gram and the following learning algorithms were evaluated: Bayesian Logistic Regression (BLR), Random Forest (RF), and Support Vector Machine (SVM). The authors also evaluated a static ensemble composed of BFR, RF and SVM using the majority vote. This was the first work to address the hate speech detection task employing an ensemble to the best of our knowledge.

%Zimmerman et al.~\cite{ZK18}, Pitsilis et al.~\cite{PRL18}, and Al-Makhadmeh and Tolba~\cite{AT20}
More recently, many works~\cite{ZK18,PRL18,PSD20,AT20,ZYH20, ASK20, ASM20, BRB20} combine deep neural networks, which is a time-consuming model compared to other algorithms such as decision trees and logistic regression. 
For instance, Paschalides et al.~\cite{PSD20} proposed a three-layer stacked ensemble classifier for hate speech detection where the deep neural networks use features from different levels: word, character, and metadata. Similarly, Montani and Schuller~\cite{MS18} also used stacking, but the meta-classifier was Logistic Regression. Regarding LR, it is worth remarking that in~\cite{PF17,PSD20}, LR obtained very competitive rates compared to the proposals.

In a different perspective, Liu et al.~\cite{HBAW19} formulated the hate speech detection task as multi-label learning where each instance can be assigned to multiple labels. For the classification phase, they proposed a fuzzy ensemble approach.

%datasets competitions 

\subsection{Literature gap}

Based on Table~\ref{tab:related_works}, we can see that more and more research works are considering an ensemble model to improve hate speech detection~\cite{ZYH20, ASM20, PSD20,  RZW19, BW15}. Some works generate an ensemble by just changing the classification model~\cite{BW15} employed, while others consider a homogeneous ensemble trained with different input features~\cite{ASK20, ZYH20}. Furthermore, multiple works only consider features from the same families such as~\cite{ASK20, ZYH20, PGLV21} which are only based on language models like BERT and other variations. As reported by Schmidt et al.~\cite{schmidt2017survey}, there exists several families of techniques that can be used either at the representation level or at the classification level for hate speech detection.

Furthermore, none of these works presents a clear methodology for analyzing multiple classification techniques and feature representation. Therefore, the field lacks a proper methodology to analyze multiple feature representation techniques and classification algorithms and use this information to optimize ensemble models. Thus, we propose a new framework to analyze the relationship between multiple techniques in order to fill this gap. The proposed framework is presented in the following section.

\section{Proposed Method}
\label{sec:approach}

The proposed framework presents a visualization tool for comparing multiple classifier models based on their prediction differences. These classifier models can be either trained using the same feature representation or a distinct one. The framework works by projecting the base models to a space called the Classifier Projection Space (CPS)~\cite{CPS2002} in which each point in this space represents a single model, and the distances between points represent the dissimilarity between distinct models (i.e., how they differ in terms of prediction). Models closer together in this space are more likely to have the same behavior, having the same prediction performance. In contrast, distant ones are more likely to behave differently (i.e., correctly classify different input samples). So they have more complementary information~\cite{featureRepresentation2013}.

\begin{figure}[htb]
	\centering
	\includegraphics[height=10cm]{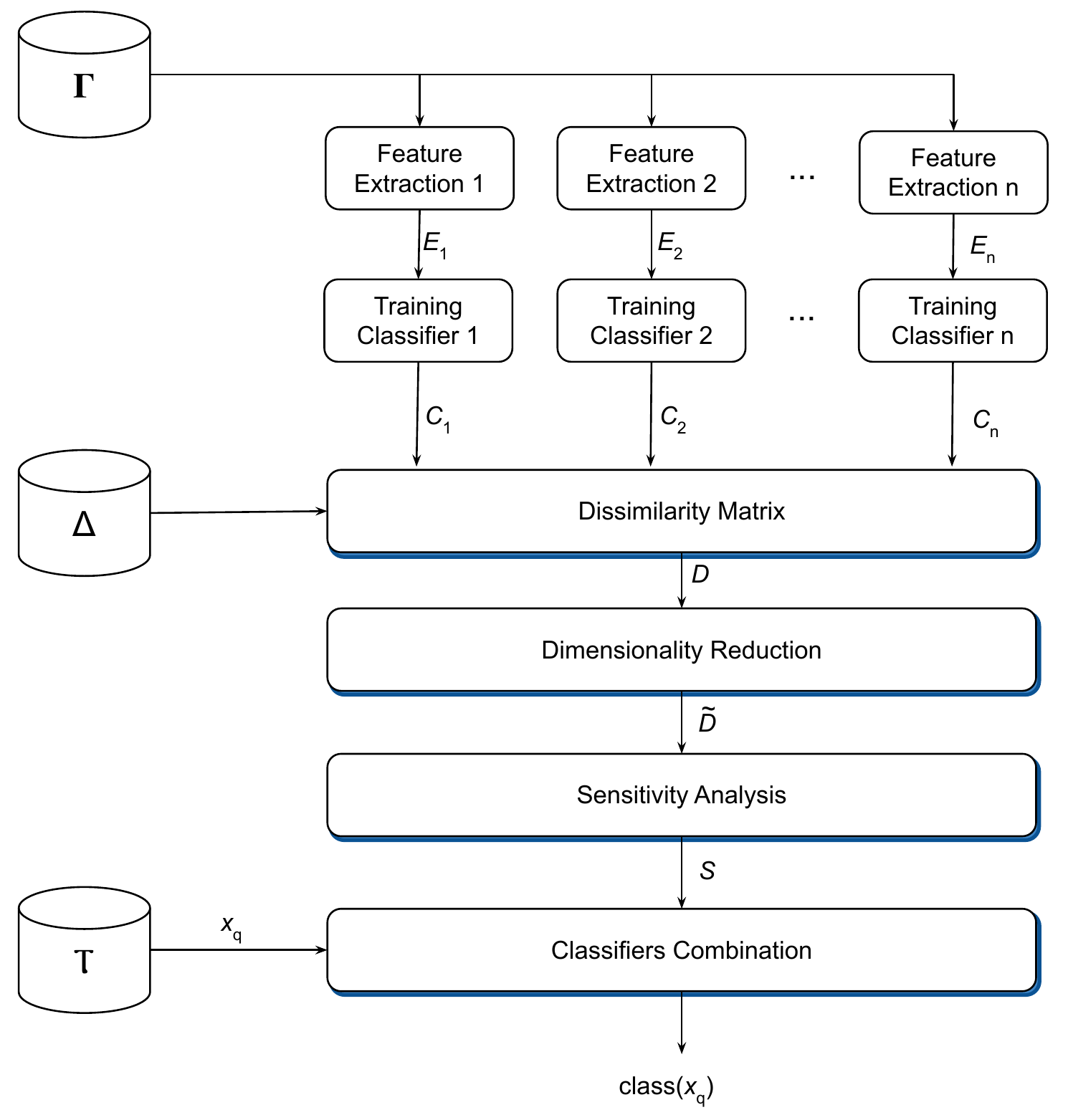}
	\caption{Proposed architecture. $\Gamma$, $\Delta$, and $\tau$ are the training, validation and testing datasets, respectively. $c_i$ is the classifier trained with the feature vector $E_i$. $\tilde{D}$ is the 2D projection of the dissimilarity matrix $D$. $S$ is a subset containing the selected classifiers. {\it class}$(x_q)$ is the predicted class of the query instance $x_q$.}
	\label{img:esquema-metodo}
\end{figure}

Figure~\ref{img:esquema-metodo} presents an overview of the proposed scheme. It is composed of 6 main steps: feature extraction, training, dissimilarity matrix, dimensionality reduction, sensitivity analysis, and classifier combination. This framework is then used to analyze the interaction between multiples feature extraction, classifier pairs and select a set containing the most complementary ones to obtain a high-performing ensemble model. 

This work considers a diverse set of classification models (e.g., SVM, Logistic Regression, Random Forest) and multiple feature representation techniques. Hence, we can have a global analysis of the relationship between state-of-the-art machine learning techniques for this application. Besides, by changing both the classifier type and feature representation, we can analyze which of these two steps contribute more to creating a complementary set of classifiers.

\subsection{Feature extraction}

In this project, five representation techniques were considered for getting a text embedding: GLoVe~\cite{pennington2014glove}, Word2Vec~\cite{mikolov2013} and fastText~\cite{fasttext-2017} representing dense word vector methods while Term-Frequency (TF) and Term-Frequency Inverse Document Frequency (TF-IDF) representing sparse feature extraction techniques based on the Bag-of-Words paradigm. At the end five sets are obtained $\{E_{GLoVe}, E_{Word2Vec}, E_{fastText}, E_{TF}, E_{TF-IDF}\}$.

Each feature extraction technique\footnote{Also referred to as feature representation or embedding.} can be seen as a different view of the problem capturing complementary information from the input text. Adopting a multiple representation approach in this scenario may be beneficial since each representation captures distinct aspects of the text. As stated by Zhao et al.~\cite{zhao2017multi}, the most significant advantage of considering multiple views of a pattern recognition problem comes from the observation that a single view cannot represent the information of all examples. With a proper combination of multiple techniques, we can increase the number of patterns represented by the system and significantly improve classification performance.					

\subsection{Training}
\par

\textcolor{blue}{Eight learning algorithms models are used in our framework: Support Vector Machines (SVM), Logistic Regression (LR), Random Forest (RF), Extra Forest (Extra), Naive Bayes (NB), K-Nearest Neighbors (KNN), Multi-Layer Perceptron Neural Network (MLP), and Convolutional Neural Network (CNN) giving a set of classifiers $C = \{C_{SVM}, C_{LR}, C_{RF}, C_{EXTRA}, C_{NB}, C_{KNN}, C_{MLP}, C_{CNN}\}$.} Each classifier $c_{i}$, from the set of classifiers $C$, is trained with a single representation of the problem $E_{j}$ extracted from the input training dataset $\Gamma$.

%Each individual classifier $c_{i}$, from a set of classifiers $C = \{c_{1}, c_{2},\ldots,c_{N}\}$, is trained with a single representation of the problem $E_{j}$ extracted from the input training dataset $\Gamma$.

After all base models $c_{i}$ are trained, their predictions $y_i$ are computed over the validation dataset $\mathcal{V}$. The use of validation data is important in order to avoid overfitting in the dissimilarity analysis. 

\subsection{Dissimilarity Matrix}

This step consists in estimating the dissimilarity in the prediction behavior from a set of classifiers $C =\{c_{1}, c_{2},\ldots,c_{N}\}$. 
In order to estimate this dissimilarity, we use diversity metrics~\cite{Kuncheva:2004:CPC:975251} which are metrics commonly used in the ensemble learning literature, as a way to analyze and generate a set of classifiers $C$ in which members present complementary behavior. Many diversity metrics are present in literature, such as classifier disagreement, Q-statistics, and the ratio of errors~\cite{doubleFault2002, aksela2006using}.

In this paper, we considered the double-fault measure proposed by Giacinto \& Roli~\cite{giacinto2001design} since it presents the highest correlation with the ensemble accuracy~\cite{doubleFault2002}. The Double Fault calculates the proportion of misclassified instances by two distinct classification models. Equation~\ref{eq:double-fault} illustrates the double fault measure.

\begin{equation}
\textit{double-fault}{(i,j)} = \frac{N^{00}}{N^{00}+N^{01}+N^{10}+N^{11}}
\label{eq:double-fault}
\end{equation}

\noindent $N^{i,j}$ is the number of instances correctly classified (1) and misclassified (0), for the $c_{i}$ and $c_{j}$, respectively. The metric equals 1 when both the classifiers always miss the same instances, make the same classification mistakes, and 0 when they never misclassify the same instance. Hence, minimizing this metric reduces the number of instances in which different classification models make the same error, achieving a more complementary classifier ensemble. We use the Double Fault implementation provided by DESlib Python library~\cite{deslib2020}.

Since the metric is higher for similar classifiers~\cite{doubleFault2002} and lower for dissimilar ones, we take its inverse to compute the dissimilarity matrix so that higher values represent more dissimilar classifiers: $d_{i, j} = \frac{1}{\textit{double-fault}(i,j)}$. After calculating $d_{i, j}$, for each pair of classifiers in the pool, we obtain an $N \times N$ matrix $D$, containing the complete diversity information between the classifiers.

\subsection{Dimensionality Reduction}

The following step is to project the matrix into a 2-dimensional representation to perform visual analysis. We use the Uniform Manifold Approximation and Projection (UMAP) proposed by McInnes et al.~\cite{mcinnes2018umap}. UMAP is a new manifold learning for dimensionality reduction technique based on Riemannian geometry~\cite{petersen2006riemannian, jost2008riemannian} that can preserve the local and the global structure of the data. Hence, it is a suitable dissimilarity reduction technique for our visualization as it preservers the distances between groups of models. 

UMAP takes as input the dissimilarity matrix $D$ and computes the $N \times 2$ matrix $\tilde{D}$. Each row of the new matrix corresponds to the position of each classifier model $c_i$. Therefore, each model $c_i$ is represented as a point in this 2-dimensional space, and the distances between points represent their dissimilarity. 

\subsection{Sensitivity Analysis}

The CPS allows us to visualize the relationship between multiple classification algorithms and identify groups of models that are either similar, represented by points close to each other in the visualization tool, or dissimilar, represented by points that are distant to each other in the CPS. Hence, this representation gives us a global view of the relationship between all models in the set of classifiers $C$.

This analysis identifies which models are more likely to improve performance by having complementary information and detecting which are redundant and should be removed from the system. Another usage of this representation is to understand how different feature representation algorithms and classification algorithms behave.

The idea is to identify groups of techniques (feature representation + classification model) close in this space and select the best performing one. Hence, we expect to reduce the computational cost by having less feature representation and classification models in the ensemble and improve classification performance as redundant models are removed in the process.

\subsection{Classifiers combination}

This module consists of combining the selected classifier's outputs to give the final prediction. Each selected model produces a vector containing the support given to each class which is aggregated to decide. There are several techniques to aggregate the outputs of multiple classifiers ranging from non-trainable, trainable, and dynamic techniques~\cite{cruz2018dynamic, Kuncheva:2004:CPC:975251}.

In this work, we propose the use of stacked generalization, which is a trainable aggregation method~\cite{wolpert1992stacked}. Stacked generalization consists of dividing the classification process into two levels of classifiers~\cite{chan1993experiments}. At the first level, the base classifiers are trained with the training dataset. These base classifiers are then used to transform the validation dataset to a vector containing the support given to each class (predictions). Those vectors are fed to a classifier in the second level, also called meta-classifier, trained to aggregate the base models' decisions into a final one. This methodology is more robust than fixed combination rules as no assumptions about the base model's behavior are required. The meta-classifier can adapt to intrinsic characteristics of the classification problem~\cite{cruz2015meta}. Thus, it is a more robust combination scheme.

It is essential to mention that the proposed framework is general enough to work with any feature extraction and classification model. New techniques can be added to the system requiring only adjustments on the dissimilarity matrix computation. The proposed methodology can continuously improve classification results as new feature representation techniques and classifiers are proposed for this application. In addition, the framework is based on comparing base models predictions in terms of diversity and requires no further information from the given task. Hence, it is application-agnostic and can be used for designing multiple classifier systems for any text classification task such as fake news detection~\cite{wang2017liar} and sentiment analysis~\cite{yadav2020sentiment}.

\section{Experimental Methodology}
\label{sec:methodology}

The experiments are conducted in order to analyze the following points:

\begin{enumerate}
	\item A comparative study considering multiple machine learning algorithm and feature representation techniques for text embedding.
	\item Analyze the relationship between multiple models, e.g., how they differ in terms of prediction performance.
	\item Propose an MCS to improve classification performance in the hate speech domains.
\end{enumerate}

\subsection{Datasets}

We considered four English hate-speech detection datasets in this study: 
\begin{enumerate}
	\item \textbf{Thomas Davidson (TD)}: was proposed by Thomas Davidson et al.~\cite{DWMW17} and is publicly available\footnote{\url{https://data.world/thomasrdavidson/hate-speech-and-offensive-language}}. This dataset consists of 24,783 labeled instances distributed over three classes: ``Hate'', ``Offensive'' and ``Non-offensive''. The dataset was built using the Twitter API to collect the sentences that contained compiled hateful terms defined on \url{hatebase.org}.
	
	\item \textbf{Zeerak Waseem (ZW)}: This dataset was proposed by Zeerak Waseem \& Dirk Hovy \cite{WH16} and is publicly available on GitHub~\footnote{\url{https://github.com/ZeerakW/hatespeech}}. The ZW dataset contains 16,907 labeled instances belonging to three classes: ``Racism'', ``Sexism'' and ``Neither''. Data was collected based on a manual search on standard terms used as insults to religious, sexual, gender, and ethnic minorities. The data was labeled by the authors first. Then it was reviewed by external evaluators to reduce the possibility of annotations being biased by either party.
	
	\item \textbf{TD + ZW}: This dataset was developed by combining the previous two datasets. We modified the labels racism and sexism of the ZW dataset to hate to unify the records, resulting in a single dataset containing 30,131 instances. We removed the records labeled as ``Neither'' since they do not have a clear class description. Since the combined datasets were proposed with different types of hate speech, their combination is expected to present multi-modal properties.
	
	\item \textbf{HatEval}: This dataset is part of the SemEval 2019 Task 5~\cite{basile2019semeval} competition which consists of hate speech detection against immigrants and women. We consider the subtask A English, a binary classification problem to detect whether a tweet in English contains hate speech. The class distribution of this dataset is almost balanced, with 42\% of the data points corresponding to a hateful text and 58\% to a non-hateful text. More information about the HatEval dataset can be found on its GitHub page: \url{https://github.com/msang/hateval}.
\end{enumerate}

We considered those datasets because they have very different characteristics in the data collection and how they address the hate speech detection problem. The TD dataset focuses on the distinction between hate, offensive, and non-offensive texts. The ZW focuses on specific types of hate speech (racism, sexism, and non-hateful). The HatEval has a distinct characteristic, having an almost balanced class distribution and presenting different types of hateful tweets (e.g., immigrants and women). Hence, by having such a diverse set of datasets, we can have a substantial evaluation of multiple feature representation and classification techniques for this task and compare the proposed method for selecting an MCS under different hate-speech detection scenarios. Furthermore, these datasets are commonly used in hate speech and offensive language detection, as illustrated in Table~\ref{tab:related_works}.

It is important to mention that all these datasets are for the English language. We decide to focus on English hate speech detection in this work as it simplifies the text preprocessing steps and the evaluation analysis. Hate speech detection for different languages such as Italian~\cite{sanguinetti2018italian} and Arabic~\cite{aljarah2020intelligent} is out of the scope of this work and will be investigated in future works.

Table~\ref{tab:dataset-segmentacao-2} summarizes all datasets and their corresponding class divisions. We conducted our experiments using a stratified 5-fold cross-validation. This methodology helps in achieving a more precise estimator of the algorithm performance as well as computing the mean and standard deviation of the results~\cite{japkowicz2011evaluating}. Furthermore, using a stratified version of cross-validation was necessary in order to avoid class bias in any of the partitions~\cite{raschka2018model}.

\begin{table}[]
	\centering
	\begin{tabular}{c|c}
		\hline
		\textbf{Dataset}   & \textbf{\#Total} \\ \hline
		Thomas Daivdson (TD)         & \begin{tabular}[c]{@{}l@{}}hateful:  1430\\ offensive: 19,190\\ non-offensive: 4,163\end{tabular} \\ \hline
		
		Zerak Wassem (ZW)            & \begin{tabular}[c]{@{}l@{}}racism: 1,970\\ sexism: 3,378\\ none: 11,559\end{tabular}              \\ \hline
		
		\multicolumn{1}{c|}{TD + ZW} & \begin{tabular}[c]{@{}l@{}}hate: 6,678\\ offensive: 19,190\\ non-offensive: 4,163\end{tabular}    \\ \hline
		
		\multicolumn{1}{c|}{HatEval} & \begin{tabular}[c]{@{}l@{}}hateful: 3783\\ non-hateful: 5271\end{tabular}                         \\ \hline
		
	\end{tabular}
	\caption{Class distribution for each dataset.}
	\label{tab:dataset-segmentacao-2}
\end{table}

For the HatEval dataset, we did not employ the 5-fold cross-validation procedure since it is originally divided into training, developing, and testing as part of the competition\footnote{Dataset used in competition can be requested in the following website: \url{http://hatespeech.di.unito.it/hateval.html}}. In this way, our results can be directly compared with the state-of-the-art from the SemEval 2019 competition.

%Each dataset was split into 67\% for training, 16.5\% for validation, and 16.5\% for the test set in a stratified fashion, i.e., maintaining the class prior probabilities in each data partition. Table~\ref{tab:dataset-segmentacao-2} shows the distribution by class for each data partition.

\subsection{Performance Metrics}

We evaluated our proposed approach using the F1-score, which is well suited for accessing machine learning models' performance under the presence of class imbalance. The F1-score is obtained by computing the harmonic mean of Precision and Recall. When dealing with multi-class classification problems, the F1-score is usually applied to each class individually and averaged using the Micro or Macro rule to give the final results. In this work, we use the Macro rule to obtain the F-Measure for the multi-class problem since it is more suited for imbalanced datasets~\cite{ferri2009experimental}. Although most hate-speech detection works use the Micro averaging scheme, this metric can mask the performance of the minority classes. As such, they are not suitable for measuring the performance of hate speech detection methods~\cite{zhang2018}.

On the other hand, the Macro averaging scheme calculates F-Measure for each label individually and then computes the results' unweighted average. Hence, all classes have the same influence on the final results despite the number of examples per class. The best method is then the one that performs well over all individual classes.

\subsection{Pre-processing and features extraction}

Before the feature extraction stage, we applied several pre-processing steps to standardize all tweets. The first step is to clean up the tweets by removing URLs, tags (i.e., "@user"), RT symbols, numbers, stop words, and redundant white spaces. We also normalized Unicode text into NFKD and converted the whole text to lowercase. 

We considered five feature extraction techniques in this study: Term Frequency (TF) and Term Frequency Inverse Document Frequency (TF-IDF) belonging to the Bag-of-Words paradigm, while Word2Vec, Global Vectors for Word Representation (GLoVe), and fastText representing dense word embeddings.

When using the Bag-of-Words (Bow) approach, the feature vector's size for each instance is equal to the vocabulary size. In contrast, the Word Embedding models have a fixed size defined in their training step. For the Word2Vec and fastText, we use the model trained over Twitter with the number of features extracted per instance was equal to 300. The GLoVe method obtained a features vector of 25 dimensions. For extraction of type BoW, we used the implementations from scikit-learn~\cite{pedregosa2011scikit}. For Word Embedding, the Zeugma library\footnote{https://zeugma.readthedocs.io/en/latest/} was used. This implementation generates an average features vector for each instance, which is then considered the text representation.

\subsection{Monolithic Classifier}
\label{sub:baseline}

We considered 8 learning algorithms in this study: Support Vector Machines (SVM)~\cite{de-gibert-etal-2018-hate,Hajime2018}, Logistic Regression~\cite{Hate-Speech-Offensive,DWMW17}, Random Forest~\cite{Hajime2018}, Naive Bayes~\cite{Hate-Speech-Offensive}, Extra Trees~\cite{van1991nlp}, K-Nearest Neighbors~\cite{ducharme2017svm}, Multi-Layer Perceptron Neural Network (MLP)~\cite{mathew2018thou,allTommi2018}, and Convolutional Neural Network (CNN)~\cite{de-gibert-etal-2018-hate, zhang2018}.
These models were selected based on the results obtained by previous work conducted in the TD, ZW and HatEval datasets. For example, the SVM classifier achieved the best performance in the HatEval competition~\cite{2019-fermi}, while Zhang et al~\cite{zhang2018} demonstrated the efficiency of CNN based models for hate speech detection using the ZW dataset. Each learning algorithm was trained with each of the feature extractors, totaling 40 (8 learning algorithm $\times$ 5 feature representation) models generated. 

%\subsubsection{Hyperparameter tuning}

\begin{table}[H]
	\centering
	\resizebox{0.5\textwidth}{!}{%
		\begin{tabular}{ll}
			\hline
			\textbf{Method} & \textbf{Hyperparameters}
			\\ \hline
			\textbf{SVM} & \begin{tabular}[c]{@{}l@{}}Kernel:[Linear, Sigmoid, rbf]\\ Gamma: [0.1, 1, 0.5]\end{tabular} \\ \hline
			\textbf{Logistic Regression} &  penalty: [l1, l2] \\ \hline
			\textbf{Random Forest} &  n\_estimators: [10. 20. 50] \\ \hline
			\textbf{Naive Bayes} & \begin{tabular}[c]{@{}l@{}}alpha: [0.1, 0.5, 1]\\ fit\_prior: [False, True]\end{tabular} \\ \hline
			\textbf{MLP} & \begin{tabular}[c]{@{}l@{}}activation: [relu, logistic]\\ solver: [adam, lbfgs]\end{tabular} \\ \hline
			\textbf{Extra Trees} & n\_estimators: [10. 20. 50] \\ \hline
			\textbf{CNN} & \begin{tabular}[c]{@{}l@{}}activation: [sigmoid, ReLU]\\ optimizer: [rmsprop, adam]\end{tabular}  \\ \hline
			\textbf{KNN} & \begin{tabular}[c]{@{}l@{}}
				%algorithm: [auto, ball\_tree, kd\_tree]\\
				n\_neighbors: [3, 5]\end{tabular} \\ \hline
		\end{tabular}%
	}
	\caption{Hyperparameter grid considered for all datasets.}
	\label{tab:hyperparameter_grid}
\end{table}

We used a grid search to find the best set of hyperparameters for each of the 40 models evaluated in this experimental study. Table~\ref{tab:hyperparameter_grid} shows the hyperparameter values considered for each classifier model. The learning algorithm was trained using the training set, while its generalization performance was measured using the validation set. For each model, the hyperparameter configuration that obtained the best performance in the validation set was selected.

For the neural network models, it was necessary to define the network architecture first. A standard architecture for the MLP was defined as consisting of a single hidden layer containing 100 neurons. For CNN,  we adopted the same 1D-CNN architecture used in previous work~\cite{zhang2018, de-gibert-etal-2018-hate}:

\begin{enumerate}
	\itemsep-0.5em 
	\item \textbf{Embedding} - Input layer containing the text embedding.
	\item \textbf{Dropout1} - Dropout layer with 20\% dropout rate.
	\item \textbf{Convolution1} - 1D convolutional layer with 64 filters, kernel size = 3, step size = 1.
	\item \textbf{Subsampling1} - 1D Max pooling operation reducing the dimensionality of the convolutional layer.
	\item \textbf{FC1} - Fully connected layer containing 256 neurons.
	\item \textbf{Dropout2} -Dropout layer with 20\% dropout rate.
	\item \textbf{Softmax} - Classification layer which produces the probabilities for each class.
\end{enumerate}

\section{Sensitivity Analysis}
\label{sec:sensitivity}

In this section, we use the proposed framework (Section~\ref{sec:approach}) to analyze the dissimilarity between different classifier types and feature representation techniques as well as define an MCS system for the hate speech detection problem. This step is conducted using the \emph{validation dataset}, $\mathcal{V}$, to prevent any bias in the final system. 

Taking into account the 40 trained classifiers (5 feature representation $\times$ 8 classification algorithms), we can consider four heuristics (groups) to compose an MCS:

\begin{itemize}
	\item \textbf{Group A} - It is a homogeneous pool composed of five classifiers that use the same classifier model (e.g., SVM), where each classifier is trained using a different feature extraction technique. The goal here is to evaluate the impact of the feature representation technique and help achieve complementary machine learning models. Eight subgroups are created using this methodology, one for each classifier model.
	
	\item \textbf{Group B} - Single feature representation technique (e.g., Word2Vec) used to generate eight classifier models. This group is evaluated to know the impact of changing the classifier model while maintaining the same feature set. Five groups (heterogeneous pools) are created with this methodology, one for each feature representation technique considered in this work.
	
	\item \textbf{Group C} - It is a heterogeneous pool composed of all 40 trained classifiers (5 feature extraction $\times$ 8 classifier model). This group aims to evaluate performance by combining all classifiers.
	
	\item \textbf{Group D} - MCS created selecting a complementary set of models by changing the feature representation and classification algorithm employed. 
	
	%	\item \textbf{Group D} - Group created using the proposed framework used to identify a set of complementary models. 
\end{itemize}

\begin{figure}[H]
	\centering
	\includegraphics[width=1\textwidth]{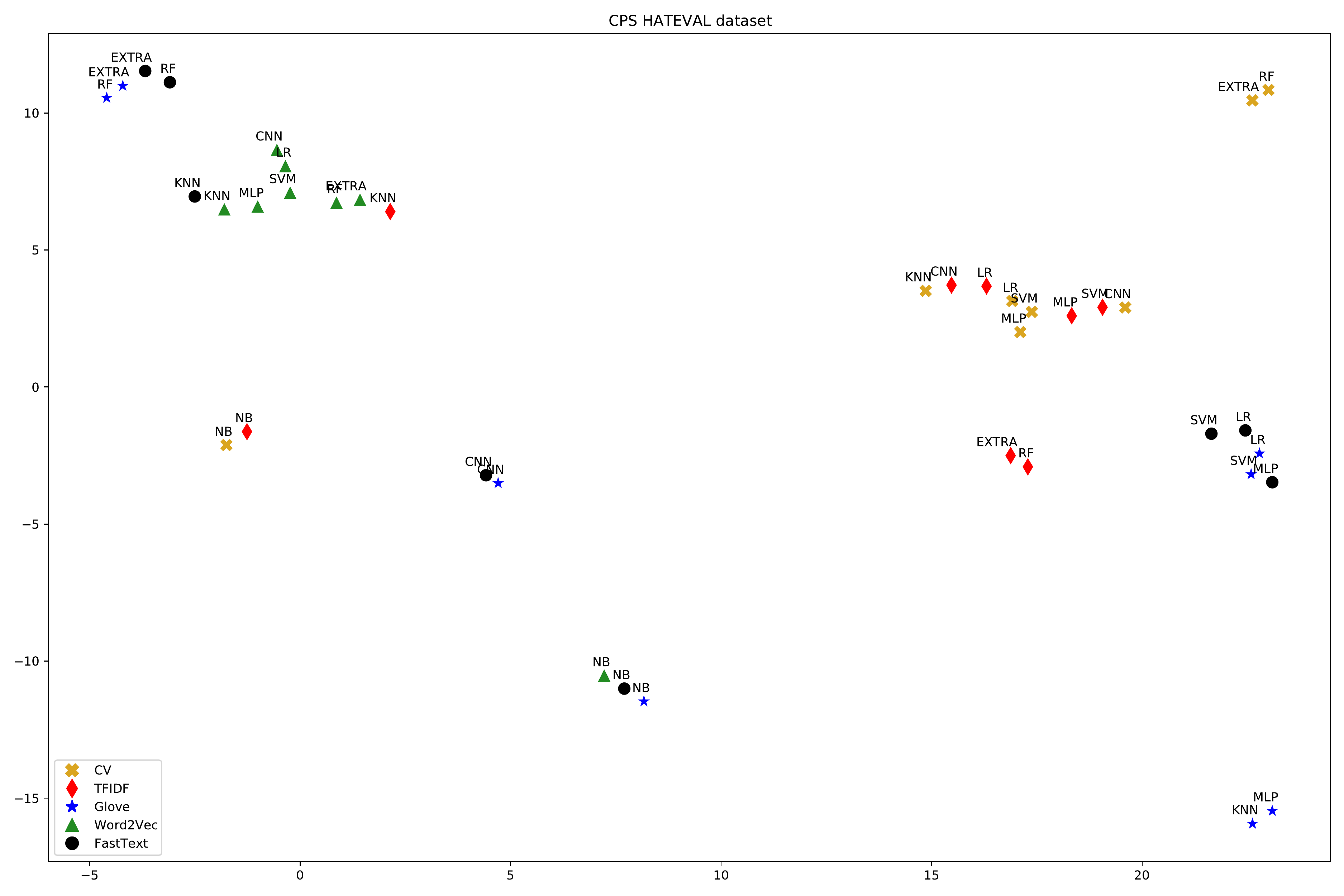}
	\caption{Classifier Projection Space (CPS) for each algorithm evaluated on the HatEval dataset.}
	\label{img:umap-hateval}
\end{figure}

Figure~\ref{img:umap-hateval} shows the CPS plot for the HatEval dataset. Based on the projection, we can easily spot a few groups of classifiers. These represent models that have the same prediction behavior and can be considered redundant. One interesting point is that many groups consist of different classifier models trained using the same feature representation. For instance, all models trained using Word2Vec as input features (represented by green triangles in Figure~\ref{img:umap-hateval}) are close together, forming a group. As such, they are likely to make the same classification mistakes.

We can also observe that, in general, we can achieve highly dissimilar models by just using a diverse set of feature representation techniques as input to train the pool of classifiers. Figure~\ref{img:umap-hateval} demonstrates that MLP classifiers having different representations as input are members of different clusters. The same behavior can be spotted for the SVM classifier. The only exception is when considering the extraction techniques from the same family, like TF and TF-IDF. In this case, the same machine learning algorithm trained with these two representation techniques is very likely to produce similar models. For instance, the models (TF $\rightarrow$ SVM) and (TF-IDF $\rightarrow$ SVM) belong to the same cluster for all the datasets evaluated in this work. The same occurs for (NB $\rightarrow$ TF) and (NB $\rightarrow$ TF-IDF), which always belong to the same cluster.

Another interesting fact is the similarity between the Random Forest and Extra Trees results which often creates a cluster. Those two classification models come from the same family (tree-based ensemble) with very little difference in how the training process is conducted~\cite{geurts2006extremely}. These observations indicate a high level of correlation between classification algorithms from the same family. In contrast, feature extraction belonging to different families (e.g., TF and GLoVe) and classification algorithms belonging to distinct families (e.g., SVM and Random Forest) are more likely to compose different clusters.

\begin{figure}[H]
	\centering
	\includegraphics[width=1\textwidth]{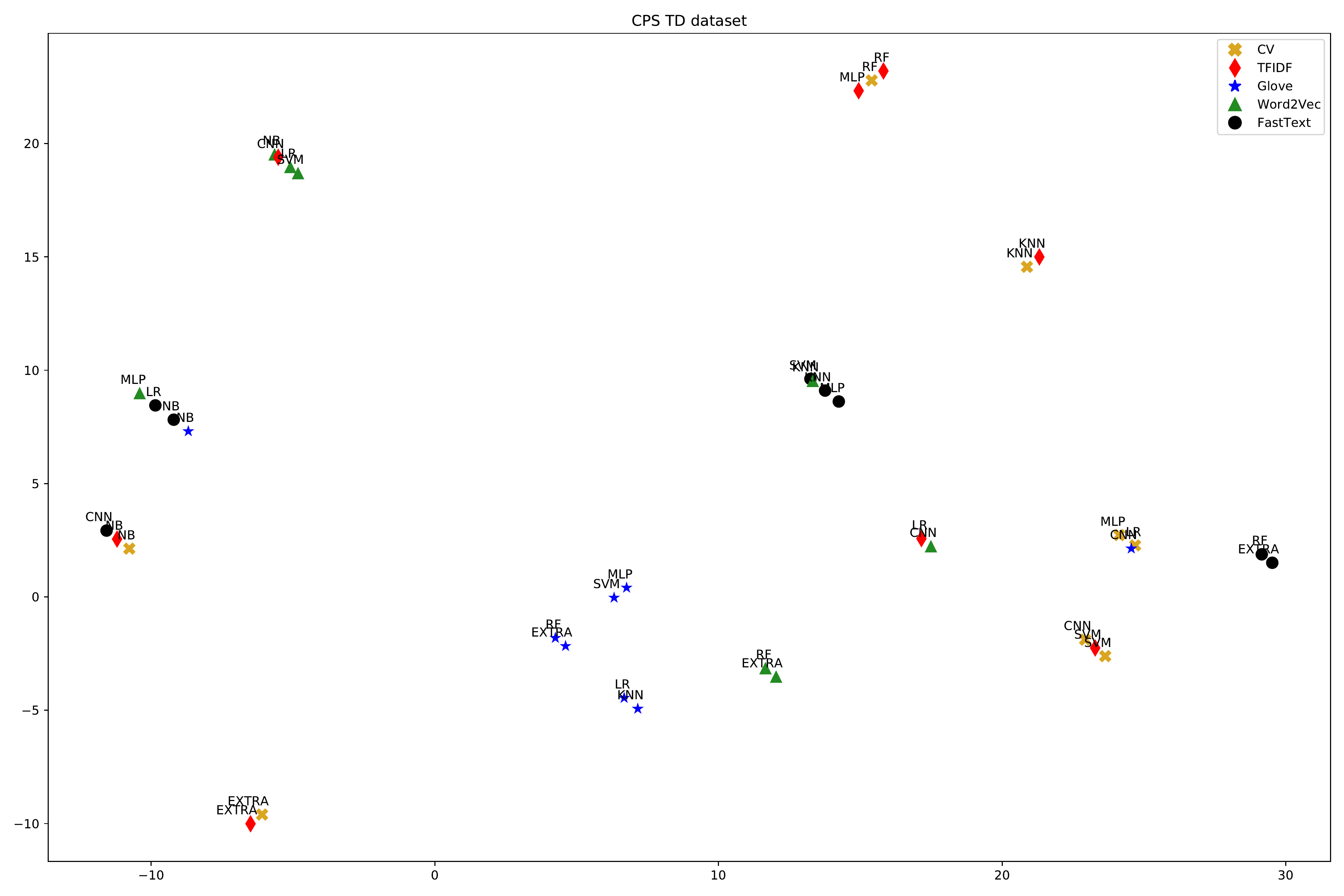}
	\caption{Classifier Projection Space (CPS) for each algorithm evaluated on the TD dataset.}
	\label{img:umap-completo-mlt-td}
\end{figure}

\begin{figure}[H]
	\centering
	\includegraphics[width=1\textwidth]{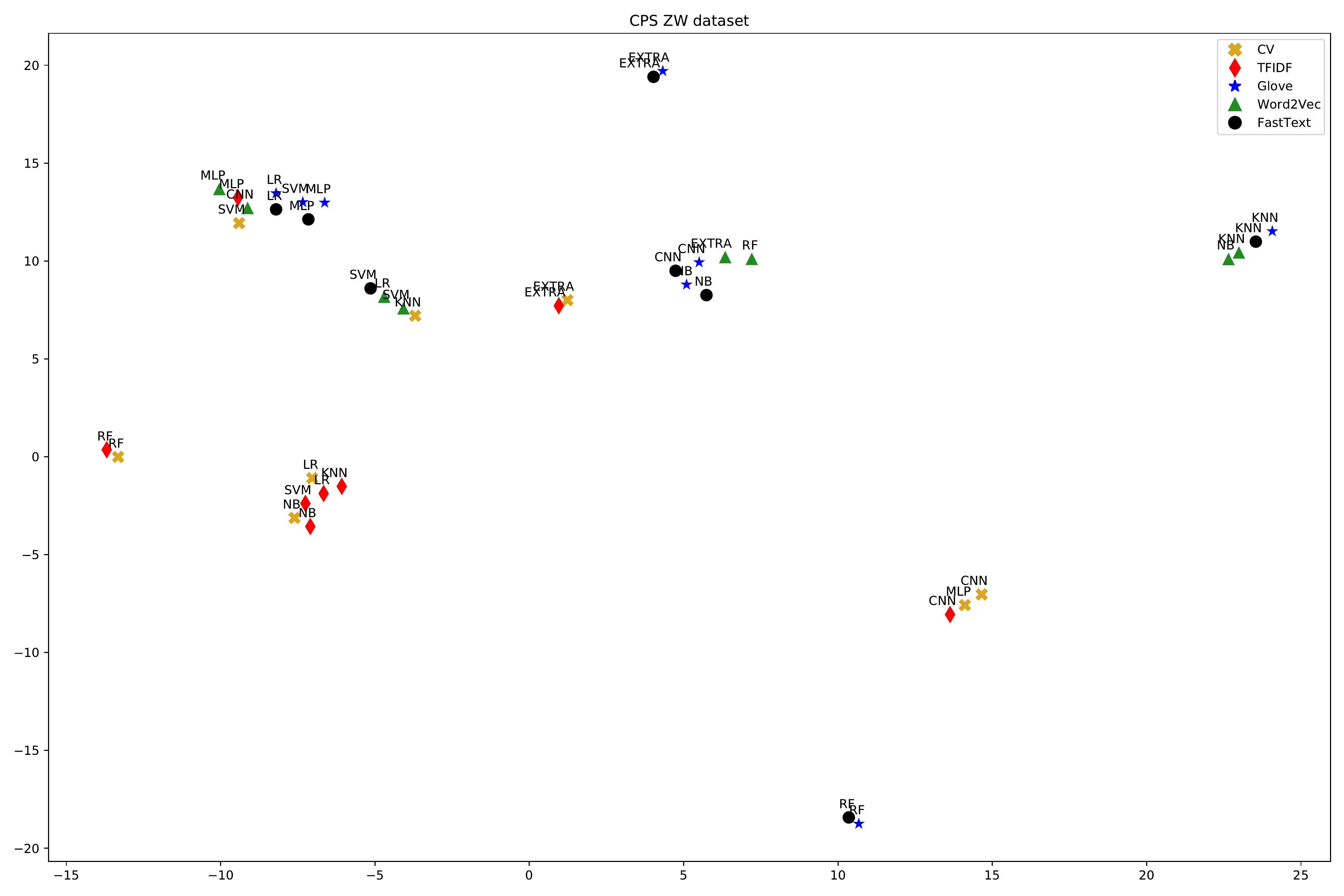}
	\caption{Classifier Projection Space (CPS) for each algorithm evaluated on the ZW dataset.}
	\label{img:umap-completo-mlt-zw}
\end{figure}

\begin{figure}[H]
	\centering
	\includegraphics[width=1\textwidth]{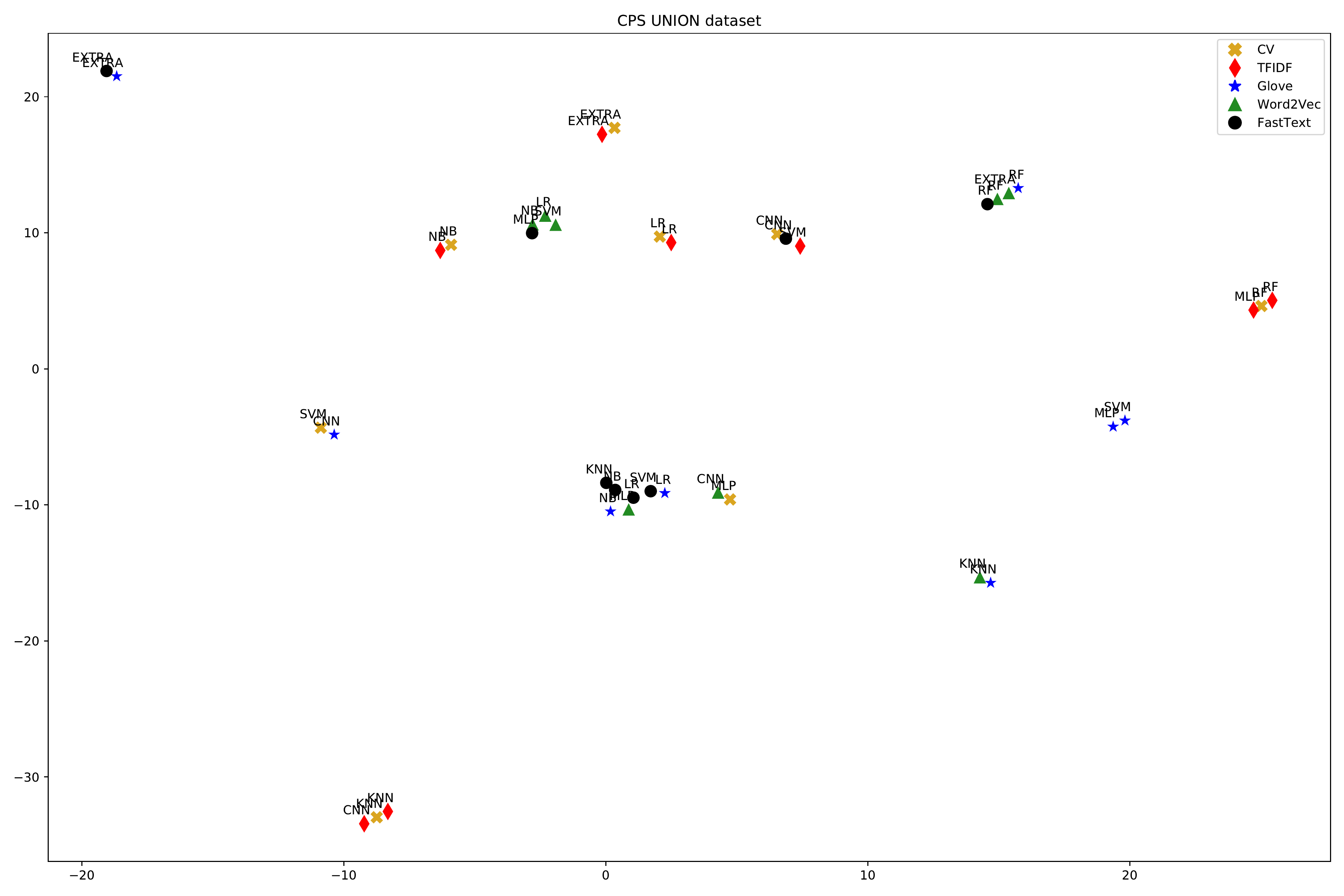}
	\caption{Classifier Projection Space (CPS) for each algorithm evaluated on the TD+ZW dataset.}
	\label{img:umap-completo-mlt-zw-td}
\end{figure}

\textcolor{blue}{Figures~\ref{img:umap-completo-mlt-td},~\ref{img:umap-completo-mlt-zw} and ~\ref{img:umap-completo-mlt-zw-td} present the CPS plot for the first cross-validation fold for the TD, ZW and TD+ZW datasets respectively. Since the CPS for each fold are consistent across multiple folds, we present only the CPS for the first cross-validation fold. CPS plot and selected models per fold are available as supplementary material on the paper GitHub page: \url{https://github.com/Menelau/Hate-Speech-MCS}}.

\textcolor{blue}{Analyzing the CPS for these three datasets, we can observe the same behavior shown in Figure~\ref{img:umap-hateval}: results of similar models such as Random Forest and Extra Trees often appear close together in the projection. The same goes for feature representation techniques similar in nature, such as TF and TF-IDF, which tend to form clusters together. The only exception is the models trained using the GloVe as feature representation which presented a more diverse behavior for TD and ZW datasets.}

\subsection{Feature representation and classifier selection}

In order to select a complementary set of classifiers by changing both the feature representation and classification models, we can use the visual information from the CPS. The goal is to avoid redundant models in the final system, as they will both decrease performance and increase the computational cost~\cite{Kuncheva:2004:CPC:975251}.

As each group represents a set of models with similar prediction behavior, only a single classifier per group can be selected to compose the final ensemble. \textcolor{blue}{Small groups composed of classifiers presenting much lower F1 score compared to the other groups (e.g., (TF $\rightarrow$ RF) and (TF $\rightarrow$ EXTRA)), were discarded from the MCS since the objective is to select a group of complementary and high performing models while having a lower quantity of selected models.} At the end, the following pairs of feature extraction and classification algorithms were selected for composing the Group D for the HatEval dataset.

\begin{itemize}
	
	\item \textbf{HatEval:} (GloVe $\rightarrow$ MLP) + (GLoVe $\rightarrow$ RF) + (GLoVe $\rightarrow$ SVM) + (W2V $\rightarrow$ NB) +  (Fast $\rightarrow$ LR) + (W2V $\rightarrow$ SVM)
\end{itemize}

Since we run a 5-fold cross-validation experiment for the TD, ZW, and TD+ZW datasets, the selected models may slightly change per fold. Sometimes, a different model belonging to the same cluster is selected among distinct folds (e.g., Extra trees instead of Random Forest). For the sake of clarity, we only present the techniques selected to compose Group D for the first fold of the experiments. They are shown below:

\begin{itemize}
	\item \textbf{ZW:} (TFIDF $\rightarrow$ EXTRA) + (W2V $\rightarrow$ NB) + (TF $\rightarrow$ NB) + (GLOVE $\rightarrow$ KNN) + (W2V $\rightarrow$ MLP) + (W2V $\rightarrow$ CNN)
	
	\item \textbf{TD:} (W2V $\rightarrow$ MLP) + (FAST $\rightarrow$ LR) + (TF $\rightarrow$ SVM) + (W2V $\rightarrow$ CNN) + (TFIDF $\rightarrow$ NB) + (W2V $\rightarrow$ NB)
	
	\item \textbf{Union:} (W2V $\rightarrow$ CNN) + (TFIDF $\rightarrow$ RF) + (W2V $\rightarrow$ NB) + (TFIDF $\rightarrow$ NB) + (TFIDF $\rightarrow$ MLP) + (W2V $\rightarrow$ MLP)
\end{itemize}

\noindent The selected models per fold are available as supplementary material on GitHub.

Nevertheless, we can observe that a combination of multiple feature representation techniques and learning algorithms are obtained at the end in all cases. Hence, the proposed visualization tool allows designing a complementary pool of classifiers containing a mix between different classification algorithms (e.g., MLP, SVM, RF, LR) and feature extraction methods (e.g., TF, W2V, GLoVe). The four groups described in this section are then used in our comparative study.

\section{Results}
\label{sec:results}

\subsection{Multiple Classifier System}
\label{sec:mcs}

Table~\ref{tab:mcs} shows the Macro F1 results for the combination techniques described in Section~\ref{sec:sensitivity} as (A), (B), (C), and (D).  Each Group A is a homogeneous pool composed of five classifiers, where each classifier is trained using a different feature extraction algorithm. For example, the first line in the table shows ``SVM(Group A)''; so, the pool in this line has five SVM, each having a different set of features. In contrast, Group B comprises eight different classifiers trained using the same feature extraction algorithms. Group C is also a heterogeneous pool, and it contains all the possible combinations between classifier and feature extraction algorithm; so, Group C has 40 classifiers (8 classifier algorithms $\times$ 5 features extraction methods). The selected classifiers using the proposed approach (described in Section~\ref{sec:approach}) composes Group D, which comprises of a mix between different set of features and classifiers selected using the visual framework.       

\begin{table}
\centering
\begin{tabular}{lcccc}
	%\toprule
	\cline{2-5}
	{} &  \multicolumn{4}{c}{Stacking LR} \\
	& TD & ZW & TD+ZW & HatEval\\
	\midrule
	SVM (Group A)                    &       0.725 & 0.753 & 0.866 & 0.500\\
	Logistic Regression (Group A)    &       0.725 & 0.760 & 0.867 & 0.478\\
	Random Forest (Group A)          &       0.635 & 0.748 & 0.851 & 0.397\\
	Naive Bayes (Group A)            &       0.653 & 0.736 & 0.830 & 0.431\\
	MLP (Group A)                    &       0.698 & 0.735 & 0.844 & 0.553 \\
	ExtraTrees (Group A)             &       0.611 & 0.710 & 0.832 & 0.408\\
	KNN (Group A)                    &       0.658 & 0.586 & 0.780 & 0.548 \\
	CNN (Group A)                    &       0.723 & 0.747 & 0.868 & 0.438\\
	
	\midrule
	TF (Group B)                     &       0.704 & 0.739 & 0.866 & 0.432 \\
	TF-IDF (Group B)                 &       0.663 & 0.765 & 0.870 & 0.397\\
	Word2Vec (Group B)               &       0.577 & 0.678 & 0.793 & 0.550\\
	GLoVe (Group B)                  &       0.561 & 0.601 & 0.773 & 0.554 \\
	fastText (Group B)               &       0.557 & 0.632 & 0.753 &  0.514\\
	\midrule
	Group C                          &       0.655 & 0.738 & 0.864 & 0.418 \\
	\midrule
	Group D &  \textbf{0.735} & \textbf{0.775} & \textbf{0.873} & \textbf{0.602} \\
	\bottomrule
\end{tabular}
\caption{Macro F1 results for four different groups of multiple classifier systems. The best result per dataset is in bold.}
\label{tab:mcs}
\end{table}

All the classifiers in the pool, independent of the Group, are combined using the same static rule: Stacking generalization~\cite{wolpert1992stacked} with the Logistic Regression as meta-classifier. The Logistic Regression used as a meta-classifier was trained to take into account the proportion of each class to tackle the class imbalance found in these datasets. Its regularization parameter (Cost) was optimized through a 5-fold cross-validation approach over the validation data using the \emph{LogisticRegressionCV} class from scikit-learn. The Macro F1 was used as the optimization metric in this process.

First, we can observe that the combination of all models (Group C) always produces inferior results than the proposed selection scheme. For instance, for the TD dataset, Group C achieved a 0.655 Macro F1-score, while Group D obtained 0.735. The same behavior happens for all datasets, showing that combining all models without applying any selection mechanism can damage the system's classification performance. Furthermore, it is more computationally expensive as it considers 40 models while the selection scheme (Group D) selects an average of six models. Thus, selection favors the overall accuracy and the computational burden.

Group C always presents inferior results compared to Groups A and B, which can be explained by the fact that either changing the feature representation or the classification algorithm leads to a set of complementary models (as illustrated by the Figure~\ref{img:umap-hateval} CPS plot). Taking ``MLP (Group A)'' for example, we can see that each MLP model, trained with a distinct feature representation, is distant in the visual analysis and belongs to different clusters. Hence, they have complementary behavior. The same conclusion can be inferred while analyzing `GLoVe (Group B)''. The vast majority of models trained with GLoVe are far apart, with only a few redundant models like Random Forests and Extra-Trees, which belong exactly to the same family. 

In all scenarios, the best result is always obtained by Group D, which consists of using the proposed framework to select a complementary set of distinct feature representations classification models. Even though changing the input features (Group A) or the classifier model (Group B) can produce a diverse ensemble, their performance is always inferior to the selection. In particular, for the HatEval dataset, the difference in performance is more evident as Group D obtained a macro F1-score of  0.602 compared to 0.554 and 0.553 of Groups A and B, respectively. This difference in results can be explained by just varying either the classifier model or the input features is not optimal, and a combination of both is required for achieving higher classification performance. In addition, there still is some redundancy in the base models, which happens when either classifier is similar in behavior, such as Random Forests and Extra Trees, or the input features are similar (e.g., TF and TF-IDF). Thus, the results confirm our hypothesis that we can build a more robust MCS by leveraging the visual tool provided by the proposed framework.

%Comparing Group A (homogeneous pool) with Group B (heterogeneous pool), the former reached better results for the TD dataset (``SVM (Group A)''), while the latter obtained better results in the other datasets using ``TF-IDF (Group B)'' for the ZW and TD+ZW datasets, and ``GLoVe (Group B)'' for the HatEval dataset. These results suggest that adopting a heterogeneous pool can be more advantageous than using a homogeneous pool, as previously reported in~\cite{wang2021hierarchical}.  

\subsection{Discussion}

Table~\ref{tab:sota} shows macro F1 results for the HatEval subtask A (English language). The proposed approach (Group D) obtained the $2^{nd}$ best place (out of 72 submissions\footnote{Ranking HatEval 2019 \url{https://docs.google.com/spreadsheets/d/1wSFKh1hvwwQIoY8_XBVkhjxacDmwXFpkshYzLx4bw-0/edit\#gid=0}}) with macro F1 equal to 0.602 (Table~\ref{tab:mcs}). Table~\ref{tab:cm} shows the confusion matrix of the proposed system that correctly classified 1010 (from 1260 hateful instances) and 938 (from 1740 non-hateful instances). 

\begin{table}
\centering
\begin{tabular}{lc}
	%\cline{2-4}
	%{} &  \multicolumn{3}{c}{Stacking LR} \\
	\hline 
	Method & Macro F1 \\ \hline 
	Fermi~\cite{2019-fermi} & 0.650 \\ \hline
	Group D$^\textbf{Proposed}$ &0.602 \\ \hline
	%Panaetius & 0.570 \\ \hline
	GLoVe (Group B)$^\textbf{Proposed}$ & 0.554 \\ \hline
	MLP (Group A)$^\textbf{Proposed}$ & 0.553 \\ \hline
	UTFPR~\cite{utfpr-2019} & 0.524\\ \hline 
	Tw-StAR~\cite{2019-twstar} & 0.503\\ \hline
	CIC-2~\cite{2019-cic} & 0.494\\ \hline
\end{tabular}
\caption{Macro F1 results for the HatEval subtask A -- English language.}
\label{tab:sota}
\end{table}

The Fermi system~\cite{2019-fermi} obtained the highest macro F1 using the Universal Sentence Encoder~\cite{cer2018universal} as features and SVM as the classifier. Keeping in mind that the proposed approach showed herein is a general framework that can accommodate many different feature extraction methods, adding new features, such as the one used in Fermi, benefits the proposal in improving its overall accuracy since more diversity is embedded into the system.

\begin{table}
\centering
\begin{tabular}{l|l|c|c|}
	\multicolumn{2}{c}{}&\multicolumn{2}{c}{Predicted label}\\
	\cline{3-4}
	\multicolumn{2}{c|}{} & Non-hateful & Hateful \\
	\cline{2-4}
	\multirow{2}{*}{True label}& Non-hateful & $802$ & $938$\\
	\cline{2-4} & Hateful & $250$ & $1010$\\
	\cline{2-4}
	%\multicolumn{1}{c}{} & \multicolumn{1}{c}{} & \multicolumn{1}{c}{$1052$} & \multicolumn{    1}{c}{$1948$} & \multicolumn{1}{c}{$3000$}\\
\end{tabular}
\caption{Confusion matrix for the Group D -- HatEval subtask A -- English language.}
\label{tab:cm}
\end{table}

The best results for Groups A and B~(Table~\ref{tab:mcs}) were obtained by the proposed ``MLP (Group A)'' and ``GLoVe (Group B)'' respectively. These two proposals are ranked $3^{rd}$ in the Ranking HatEval 2019. Such results show that combining different feature extraction methods and classifiers is a promising alternative to deliver robust and accurate results.  
Logistic Regression (LR) was the meta-classifier employed in the experiments. In LR, each input variable affects the target output that is expressed by the variable’s coefficient. In other words, these coefficients could help to understand the importance of each input variable to the meta-classifier decision. The input of our meta-classifier is the output of the trained models. Given that all classifiers’ outputs are in the same range, this standardization helps make the coefficient scale comparable with each other. For instance, the proposed approach (Group D) selected 6 out of 40 pairs of classifier and feature extraction algorithms, which are: MLP-GloVe, RF-GLoVe, SVM-GLoVe, NB-W2V, LR-Fast, and SVM-W2V. These pairs are ordered from the one having the highest coefficient ($\approx 1.06$ for MLP-GLoVe) to the smallest ($\approx 0.03$ for SVM-W2V) one. In other words, MLP-GLoVe plays a more important role than SVM-W2V in the meta-classifier prediction.

HatEval makes available a clear experimental protocol where the dataset is already split. Thus, the comparison between the proposal and the literature is simplified. Unfortunately, this is not the case for the other datasets evaluated in this paper. There is no standard methodology for the TD and ZW datasets, and literature works use many different training/testing split strategies and measures. Besides, the source code is rarely publicly available. So, the comparison with the literature methods is quite challenging in these conditions. This paper fulfills this gap by publicly making all the splits per dataset and the source code of the methods used herein. In this vein, the community can replicate the results and use the repository to further extend the area's knowledge.

\section{Conclusion}
\label{sec:conclusion}

Automating hate speech detection in social networks is a difficult task due to the high volume of data generated daily, and the definition of what is considered hate speech is still much discussed. This work presented an empirical evaluation of multiple feature extraction techniques and classification models for detecting hate speech in social networks.
We adopted five different feature extraction methods and eight classification algorithms, giving a total of 40 hate-speech classification models in our experimental study. Then we propose a framework based on diversity metrics in the classifier projection space to understand the relationship between them.

The proposed framework is used to analyze the behavior of four heuristics to design an MCS for hate-speech detection:
\begin{itemize}
	\item Same classifier model trained with different feature representation;
	\item Multiple classifier models having the same feature representation as input features;
	\item A combination of all classifiers generated;
	\item An MCS containing the most complementary set of classifiers selected using the proposed framework.
\end{itemize}

Experimental results considering four hate-speech classification datasets show that combining all models (Group C) hinders classification performance compared to any of the selection heuristics evaluated in this work. We also observed that changing the feature representation and using classification algorithms from different families (e.g., SVM, Random Forest, Naive Bayes) can lead to good MCS as they generate complementary models according to the proposed framework for analyzing their relationship. Furthermore, using the proposed framework as a guide for selecting a complementary set of models mixing the classification algorithm and the input feature representation achieved the best results for the four datasets considered in this study. 

We also compared the performance of the proposed combination schemes against state-of-the-art techniques. The selection of feature representation and classification methods using the proposed framework (Group D) is ranked second in the Hateval 2019 competition. Furthermore, we also observed that Group B and Group A are ranked third and fourth, respectively, in the competition. Such results show that combining a complementary set of feature extraction methods and classification algorithms is a promising alternative to deliver robust and accurate results for hate speech detection. Thus, we believe the proposed framework is a promising alternative for analyzing and designing better MCS. 

It is essential to mention that the proposed framework for feature representation and classifier selection works just by comparing the predictions of different pairs (feature representation, classifier) and requires no specific information about the given classification task. Hence, the framework is application-agnostic and can be applied for different text classification tasks such as fake news classification~\cite{wang2017liar} as well as for detecting hate speech in different languages such as Spanish~\cite{basile2019semeval}, Arabic~\cite{aljarah2020intelligent}, and Italian~\cite{sanguinetti2018italian}. We will investigate these points in future works.

\bibliographystyle{elsarticle-num}
\bibliography{main}

\end{document}